\documentclass[10pt,twocolumn,letterpaper]{article}

\usepackage{iccv}
\usepackage{times}
\usepackage{epsfig}
\usepackage{graphicx}
\usepackage{amsmath}
\usepackage{amssymb}
\usepackage{xcolor}
\usepackage{comment}
\usepackage{multirow}
\usepackage{float}
\usepackage[caption = false]{subfig}
\usepackage{adjustbox}
\usepackage[none]{hyphenat}
\usepackage{authblk}
\usepackage{array}
\newcolumntype{P}[1]{>{\centering\arraybackslash}p{#1}}
\usepackage{caption}

\usepackage{algorithm}
\usepackage{algorithmic}

\usepackage[pagebackref=true,breaklinks=true,letterpaper=true,colorlinks,bookmarks=false]{hyperref}

\DeclareMathOperator{\DROP}{DROP}
\DeclareMathOperator{\EDA}{EDA}
\DeclareMathOperator{\LN}{LN}
\DeclareMathOperator{\MLP}{MLP}

\iccvfinalcopy 

\def\iccvPaperID{1840} 

\ificcvfinal\fi

\begin{document}



\title{When Pigs Fly: Contextual Reasoning in Synthetic and Natural Scenes}


\author[1,*]{\vspace{-0.6cm} Philipp Bomatter}
\author[2,3,*]{Mengmi Zhang}
\author[4]{Dimitar Karev}
\author[3,5]{Spandan Madan}
\author[4]{\par Claire Tseng}
\author[2,3]{Gabriel Kreiman}

\affil[1]{\small ETH Z\"urich}
\affil[2]{Children's Hospital, Harvard Medical School}
\affil[3]{Center for Brains, Minds and Machines}
\affil[4]{Harvard College, Harvard University}
\affil[5]{School of Engineering and Applied Sciences, Harvard University}
\affil[*]{Equal contribution}
\affil[ ]{\small \par Address correspondence to gabriel.kreiman@tch.harvard.edu}

\twocolumn[{%
\renewcommand\twocolumn[1][]{#1}%
\maketitle

\vspace{-1cm}
\begin{center}
    \centering
    \includegraphics[width=17cm, height = 5.0cm]{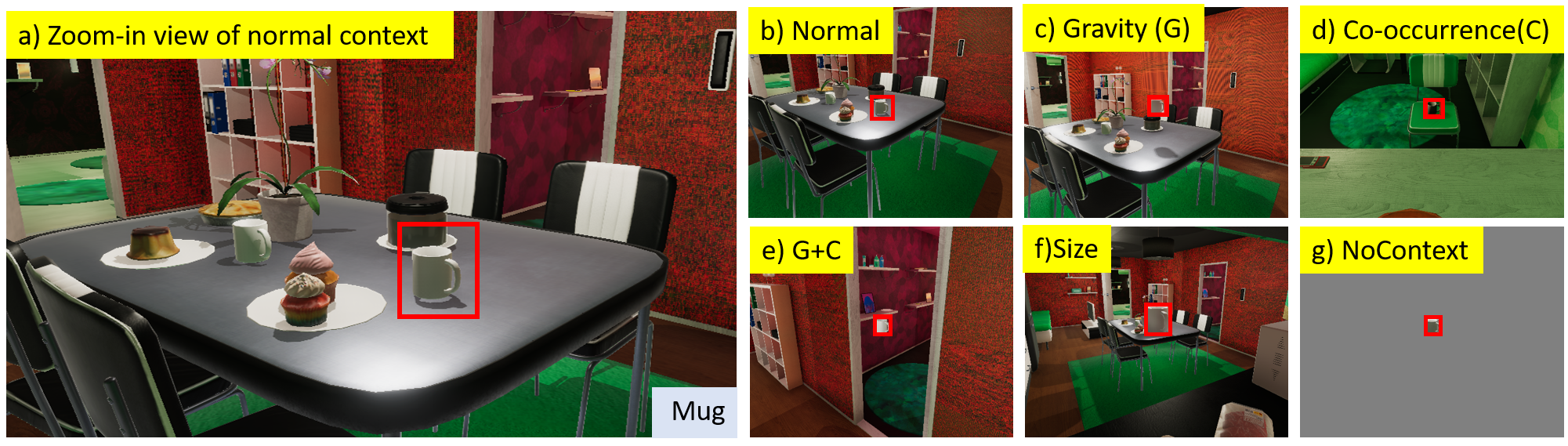}\vspace{-4mm}
    \captionof{figure}{\textbf{Images under normal context and out-of-context conditions were generated in the VirtualHome environment \cite{puig2018virtualhome} using the Unity 3D simulation engine \cite{juliani2018unity}.} 
    The same target object (a mug, red bounding box) is shown in different context conditions: normal context (a, b) and out-of-context conditions including gravity ((c), target object is floating in the air), change in object co-occurrence statistics (d), combination of both gravity and object co-occurrence statistics (e), enlarged object size (f), and no context with uniform grey pixels as background (g).
    }\label{fig:fig1}
\end{center}%
\vspace{0.2cm}
}]

\ificcvfinal\thispagestyle{empty}\fi

\begin{abstract}
    Context is of fundamental importance to both human and machine vision; e.g., an object in the air is more likely to be an airplane than a pig. The rich notion of context incorporates several aspects including physics rules, statistical co-occurrences, and relative object sizes, among others. While previous work has focused on crowd-sourced out-of-context photographs from the web to study scene context, controlling the nature and extent of contextual violations has been a daunting task. Here we introduce a diverse, synthetic \textbf{O}ut-of-\textbf{C}ontext \textbf{D}ataset (OCD) with fine-grained control over scene context. By leveraging a 3D simulation engine, we systematically control the gravity, object co-occurrences and relative sizes across 36 object categories in a virtual household environment. We conducted a series of experiments to gain insights into the impact of contextual cues on both human and
    machine vision using OCD. We conducted psychophysics experiments to establish a human benchmark for out-of-context recognition, and then compared it with state-of-the-art computer vision models to quantify the gap between the two. We propose a context-aware recognition transformer model, fusing object and contextual information via multi-head attention. Our model captures useful information for contextual reasoning, enabling human-level performance and better robustness in out-of-context conditions compared to baseline models across OCD and other out-of-context datasets. All source code and data are publicly available at \url{https://github.com/kreimanlab/WhenPigsFlyContext}
\end{abstract}

\section{Introduction}
A coffee mug is usually a small object (Fig.\ref{fig:fig1}a), which
does not fly on its own (Fig.\ref{fig:fig1}c) and can often be found on a table (Fig.\ref{fig:fig1}a) but not on a chair (Fig.\ref{fig:fig1}d). Such contextual cues have a pronounced impact on the object recognition capabilities of both humans \cite{zhang2020putting}, and computer vision models \cite{torralba2003context, choi2012context, oliva2007role, mack2011object}.
Neural networks learn co-occurrence statistics between an object's appearance and its label, but also between the object's context and its label \cite{divvala2009empirical, sun2017seeing, beery2018recognition}. Therefore, it is not surprising that recognition models fail to recognize objects in unfamiliar contexts \cite{rosenfeld2018elephant}. 
Despite the fundamental role of context in visual recognition, it remains unclear \emph{what}  contextual cues should be integrated with object information and \emph{how}.

Two challenges have hindered progress in the study of the role of contextual cues: (1) context has usually been treated as a monolithic concept and (2) 
large-scale, internet-scraped datasets like ImageNet \cite{deng2009imagenet} or COCO \cite{lin2014microsoft}
are highly uncontrolled.
To address these challenges, we present a methodology to systematically study the effects of an object's context on recognition by leveraging a Unity-based 3D simulation engine for image generation \cite{juliani2018unity}, and manipulating 3D objects in a virtual home environment \cite{puig2018virtualhome}. The ability to rigorously control every aspect of the scene enables us to systematically violate contextual rules and assess their impact on recognition. 
We focus on three fundamental aspects of context:  (1) \emph{gravity} - objects without physical support, (2) \emph{object co-occurrences} - unlikely object combinations, and (3) \emph{relative size} - changes to the size of target objects relative to the background. As a critical benchmark, we conducted psychophysics experiments to measure human performance and compare it with state-of-the-art computer vision models.

We propose a new context-aware architecture, which can incorporate object and contextual information to achieve higher object recognition accuracy given proper context and robustness to out-of-context situations.
Our \textbf{C}ontext-aware \textbf{R}ecognition \textbf{T}ransformer  \textbf{N}etwork (\textit{CRTNet}) uses two separate streams to process the object and its context independently before integrating them via multi-head attention in transformer decoder modules. Across multiple datasets, the CRTNet model surpasses other state-of-the-art computational models in normal context and classifies objects robustly despite large contextual variations, much like humans do.

Our contributions in this paper are three-fold. Firstly, we introduce a challenging new dataset for in- and out-of-context object recognition that allows fine-grained control over context violations including gravity, object co-occurrences and relative object sizes (\emph{out-of-context dataset}, OCD). Secondly, we conduct psychophysics experiments to establish a human benchmark for in- and out-of-context recognition and compare it with state-of-the-art computer vision models.
Finally, we propose a new context-aware architecture for object recognition, which combines object and scene information to reason about context and generalizes well to out-of-context images. We  release  the  entire  dataset,  including  our  tools  for  the generation of additional images and the source code for CRTNet at \href{https://github.com/kreimanlab/WhenPigsFlyContext}{https://github.com/kreimanlab/WhenPigsFlyContext}.


\section{Related Works}
    \textbf{Out-of-context datasets:}
        Notable works on out-of-context datasets include the UnRel dataset \cite{Peyre17} and the Cut-and-paste dataset presented in \cite{zhang2020putting}. While UnRel is a remarkable collection of out-of-context natural images, it is limited in size and diversity. 
        A drawback of cutting-and-pasting \cite{ghiasi2020simple} 
        is the introduction of artifacts such as unnatural lighting, object boundaries, sizes and positions. 
        Neither of those datasets 
        allow systematic analysis of individual properties of context. 
        3D simulation engines enable easily synthesizing many images and systematically investigating the violation of contextual cues. It is challenging to achieve these goals with real-world photographs. Moreover, these simulation engines enable 
        precise control of contextual parameters, changing cues one at a time in a systematic and quantifiable manner.




\textbf{Out-of-context object recognition:}
In previous work, context has mostly been studied as a monolithic property in the form of the target object's background. Previous work included testing the generalization to new backgrounds \cite{beery2018recognition} and incongruent backgrounds \cite{zhang2020putting}, exploring the impact of foreground-background relationships on data augmentation \cite{dvornik2018modeling}, and replacing image sub-regions by another sub-image, i.e. object transplanting \cite{rosenfeld2018elephant}.
In this paper, we evaluate different properties of contextual cues (\eg gravity) in a quantitative, controlled, and systematic manner.


\textbf{3D simulation engines and computer vision:} Recent studies have demonstrated the success of using 3D virtual environments for tasks such as object recognition with simple and uniform backgrounds \cite{borji2016ilab}, routine program synthesis \cite{puig2018virtualhome}, 3D animal pose estimation \cite{mu2020learning}, and studying the generalization capabilities of CNNs \cite{madan2020capability, halder2019physics}. However, to the best of our knowledge, none of these studies have tackled the challenging problem of how to integrate contextual cues.






\textbf{Models for context-aware object recognition:} To tackle the problem of context-aware object recognition, researchers have proposed classical approaches, \eg Conditional Random Field (CRF) \cite{gonfaus2010harmony,yao2012describing,ladicky2010graph,chen2018deeplab}, and graph-based methods \cite{torralba2003contextual,wu2018learning,torralba2005contextual, choi2012context}.  Recent studies have extended this line of work to deep graph neural networks \cite{hu2016learning,choi2012unified,deng2016structure,battaglia2016interaction}.
Breaking away from these previous works where graph optimization is performed globally for contextual reasoning in object recognition, our model has a two-stream architecture which separately processes visual information on both target objects and context, and then integrates them with multi-head attention in stacks of transformer decoder layers. In contrast to other vision transformer models in object recognition \cite{dosovitskiy2020image} and detection \cite{carion2020end}, CRTNet performs in-context recognition tasks given the target object location.


\begin{figure*}[ht]
    \begin{center}
        \includegraphics[width=16.5cm]{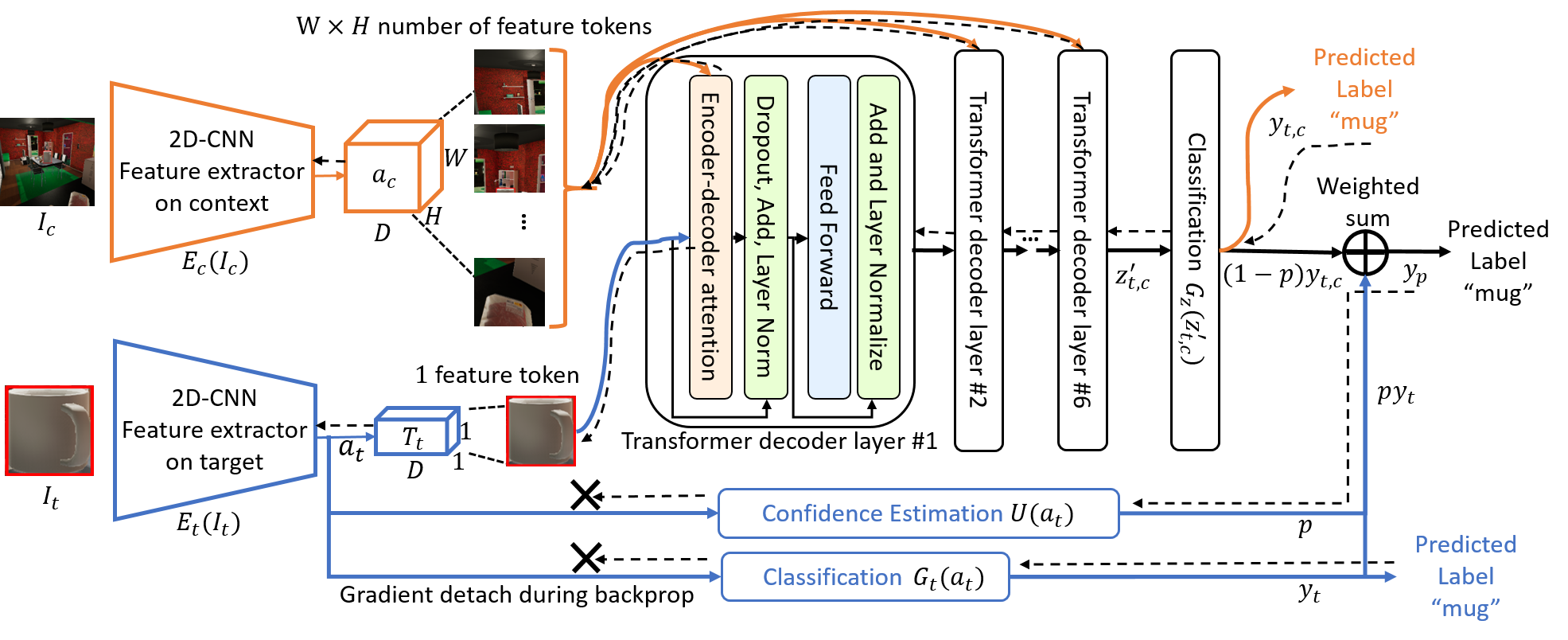}\vspace{-8mm}
    \end{center}
    
    \caption{
    \textbf{Architecture overview of the Context-aware Recognition Transformer Network (CRTNet).}
    CRTNet consists of 3 main modules: feature extraction, integration of context and target information, and confidence-modulated classification. CRTNet takes the cropped target object $I_t$ and the entire context image $I_c$ as inputs and extracts their respective features. These feature maps are then tokenized and the information of the two streams is integrated over multiple transformer decoder layers. CRTNet also estimates a confidence score for recognizing the target object based on object features alone, which is used to modulate the contributions of $y_t$ and $y_{t,c}$ to the final prediction $y_p$. The dashed lines in backward direction denote gradient flows during backpropagation. The two black crosses denote where the gradient updates stop.
    See Sec. \ref{sec:model} for details.
    }
    \vspace{-5mm}\label{fig:fig3}
\end{figure*}

\section{Context-aware Recognition Transformer}\label{sec:model}

\subsection{Overview}


We propose the Context-aware Recognition Transformer Network (CRTNet, Figure \ref{fig:fig3}). CRTNet is presented with an image with multiple objects and a bounding box to indicate the target object location. 
The model has three main elements:
First, CRTNet uses a stack of transformer decoder modules with multi-head attention to hierarchically reason about context and integrate contextual cues with object information. Second, a confidence-weighting mechanism  improves the model's robustness and gives it the flexibility to select what information to rely on for recognition. 
Third, we curated the training methodology with gradient detachment to prioritize important model components and ensure efficient training of the entire architecture.

Inspired by the eccentricity dependence of human vision, CRTNet has one stream that processes only the target object ($I_t, 224 \times 224$), and a second stream 
devoted to the periphery ($I_c, 224 \times 224$). $I_t$ is obtained by cropping the input image to the bounding box
whereas $I_c$ covers the entire contextual area of the image.
$I_c$ and $I_t$ are resized to the same dimensions. Thus, the target object's resolution is higher in $I_t$. The two streams are encoded through separate 2D-CNNs. After the encoding stage, CRTNet tokenizes the feature maps of $I_t$ and $I_c$, integrates object and context information via hierarchical reasoning through a stack of transformer decoder layers, and predicts class label probabilities $y_{t,c}$ within $C$ classes.


A model that always relies on context can make mistakes under unusual context conditions. To increase robustness, CRTNet makes a second prediction $y_t$, based on target object information alone, estimates the confidence $p$ of this prediction, and computes a confidence-weighted average of $y_t$ and $y_{t,c}$ to get the final prediction $y_p$. 
If the model makes a confident prediction based on the target object alone, this decision overrules the context reasoning stage.

\subsection{Convolutional Feature Extraction}

CRTNet takes $I_c$ and $I_t$ as inputs and uses two 2D-CNNs, $E_c(\cdot)$ and $E_t(\cdot)$, to extract context and target feature maps $a_c$ and $a_t$, respectively, where $E_c(\cdot)$ and $E_t(\cdot)$ are parameterized by $\theta_{E_c}$ and $\theta_{E_t}$.
We use the DenseNet architecture \cite{huang2017densely} with weights pre-trained on ImageNet \cite{deng2009imagenet} and fine-tune it.
Assuming that different features in $I_c$ and $I_t$ are useful for recognition, we do not enforce sharing of the parameters $\theta_{E_c}$ and $\theta_{E_t}$. We demonstrate the advantage of non-shared parameters in the ablation study (Sec. \ref{subsec:ablation}). To allow CRTNet to focus on specific parts of the image and select features at those locations, we preserve the spatial organization of features
and define $a_c$ and $a_t$ 
as the output feature maps from the last convolution layer of DenseNet. Both $a_c$ and $a_t$ are of size $D \times W \times H = 1,664 \times 7 \times 7$, where 
$D$, $W$ and $H$ denote the number of channels, width and height of the feature maps respectively.





\subsection{Tokenization and Positional Encoding}\label{sec:tokenization}


We tokenize the context feature map $a_c$ by splitting it into patches based on locations, following \cite{dosovitskiy2020image}. Each context token corresponds to a feature vector $\mathbf{a_c^{i}}$ of dimension $D$ at location $i$ where $i\in \{1,..,L=H \times W\}$.
To compute target token $T_t$, CRTNet aggregates the target feature map $a_t$ via average pooling:


\vspace{-2mm}
\begin{equation}
T_t = \frac{1}{L} \sum_{i=1,...,L} \mathbf{a_t^i}
\end{equation}


To encode the spatial relations between the target token and the context tokens, as well as between different context tokens, we learn a positional embedding of size $D$ for each location $i$ and add it to the corresponding context token $\mathbf{a_c^i}$. For the target token $T_t$, we use the positional embedding corresponding to the location, within which the bounding box midpoint is contained. The positionally-encoded context and target tokens are denoted by $z_c$ and $z_t$ respectively.




\subsection{Transformer Decoder}
We follow the original transformer decoder 
\cite{vaswani2017attention}, taking $z_c$ to compute keys and values, 
and $z_t$ to generate the queries in the transformer encoder-decoder multi-head attention layer. 
Since we only have a single target token, we omit the self-attention layer.
In the experiments, we also tested CRTNet with self-attention enabled and we did not observe performance improvements.
 Our decoder layer consists of alternating layers of encoder-decoder attention (EDA) and multi-layer perceptron (MLP) blocks. Layernorm (LN) is applied after each residual connection. Dropout (DROP) is applied within each residual connection and MLP block. The MLP contains two layers with a ReLU non-linearity and DROP.

\vspace{-2mm}
\begin{equation}\label{equ:equEDA}
z_{t,c} = \LN(\DROP(\EDA(z_t, z_c)) + z_t)
\end{equation}
\begin{equation}\label{equ:equMLP}
z'_{t,c} = \LN(\DROP(\MLP(z_{t,c})) + z_{t,c})
\end{equation}

Our transformer decoder has a stack of $X=6$ layers, indexed by $x$. We repeat the operations in Eqs~\ref{equ:equEDA} and~\ref{equ:equMLP} for each transformer decoder layer by recursively assigning $z'_{t,c}$ back to $z_t$ as input to the next transformer decoder layer. Each EDA layer integrates useful information from the context and the target object with 8-head selective attention
. Based on accumulated information from all previous $x-1$ layers, each EDA layer enables CRTNet to progressively reason about context
by updating the attention map on 
$z_c$ over all $L$ locations. 
We provide visualization examples of attention maps along the hierarchy of the transformer decoder modules in Supp. Fig S1.


\subsection{Confidence-modulated Recognition}
The context classifier $G_z(\cdot)$ with 
parameters $\theta_{G_z}$ consists of a fully-connected layer and a softmax layer. It takes the feature embedding $z'_{t,c}$ from the last transformer decoder layer and outputs the predicted class distribution vector: $y_{t,c} = G_z(z'_{t,c})$.
Similarly, the target classifier $G_t(\cdot)$, takes the feature map $a_t$ as input and outputs the predicted class distribution vector: $y_t = G_t(a_t)$.

Since neural networks are often fooled by incongruent context \cite{zhang2020putting}, we propose a confidence-modulated recognition mechanism balancing the predictions from $G_t(\cdot)$ and $G_z(\cdot)$.
The confidence estimator $U(\cdot)$ with parameters $\theta_U$
takes the target feature map $a_t$ 
as input and outputs a value $p$ indicating how confident CRTNet is about the prediction
$y_t$. $U(\cdot)$ is a feed-forward multi-layer perceptron network with a sigmoid function to normalize the confidence score to [0, 1].
\begin{equation}\label{equ:confidencemod}
p = \frac{1}{1+e^{-U(a_t)}}
\end{equation}
We use $p$ to compute a confidence-weighted average of $y_{t,c}$ and $y_t$ for the final predicted class distribution $y_p$: 
$y_p =  p y_t + (1-p) y_{t,c}$.
The higher the confidence $p$, the more CRTNet relies on the target object itself, rather than on the integrated contextual information, for classification. We demonstrate the advantage of using $y_p$ rather than $y_{t,c}$ or $y_t$ as a final prediction in the ablation study (Sec. \ref{subsec:ablation}).


\subsection{Training}\label{subsec:trainingtest}

CRTNet is trained end-to-end with three loss functions:
(i) to train the confidence estimator $U(\cdot)$, we use a cross-entropy loss with respect to the confidence-weighted prediction $y_p$. This allows $U(\cdot)$ to learn to increase the confidence value $p$ when the prediction $y_t$ based on target object information alone is correct. (ii) To train $G_t(\cdot)$, we use a cross-entropy loss with respect to $y_t$. (iii) For the other components of CRTNet, including the transformer decoder modules and the classifier $G_z(\cdot)$, we use a cross-entropy loss with respect to $y_{t,c}$. Instead of training everything based on $y_p$, the three loss functions together maintain strong learning signals for all parts in the architecture irrespective of the confidence value $p$.

To facilitate learning for specific components in CRTNet, we also introduce gradient detachments during backpropagation (Fig.~\ref{fig:fig3}). 
Gradients flowing through both $U(\cdot)$ and $G_t(\cdot)$ are detached from $E_t(\cdot)$ to prevent them from driving the target encoder to learn more discriminative features, which could impact the efficacy of the transformer modules and $G_z(\cdot)$. 
We demonstrate the benefit of these design decisions in ablation studies (Sec. \ref{subsec:ablation}).

\section{Experimental Details}\label{sec:dataset}
\subsection{Baselines}\label{subsec:baselines}

\textbf{CATNet \cite{zhang2020putting}} is a context-aware two-stream object recognition model. It processes the visual features of a cropped target object and context in parallel, dynamically incorporates object and contextual information by constantly updating its attention over image locations, and sequentially reasons about the class label for the target object via a recurrent neural network.  

\textbf{Faster R-CNN \cite{ren2015faster}} is an object detection algorithm. We adapted it to the context-aware object recognition task by replacing the region proposal network with the ground truth bounding box indicating the location of the target object. 

\textbf{DenseNet \cite{huang2017densely}} is a 2D-CNN  with dense connections that takes the cropped target object patch $I_t$ as input.


\begin{figure*}[t]
\centering
\subfloat[OCD]{\includegraphics[width= 2.5cm]{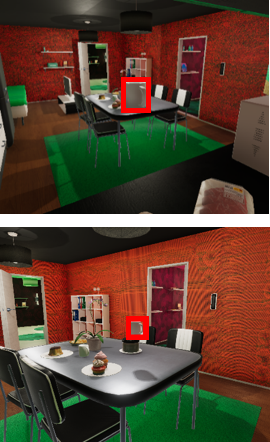}\label{fig:fig2a}}\hspace{0.2cm}
\subfloat[Cut-and-paste]{\includegraphics[width= 2.5cm]{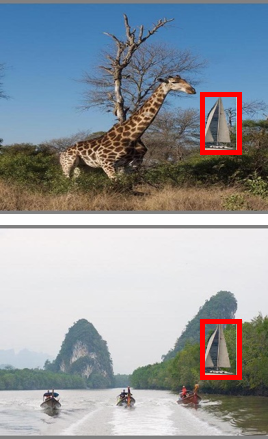}\label{fig:fig2b}}\hspace{0.2cm}
\subfloat[UnRel]{\includegraphics[width= 2.5cm]{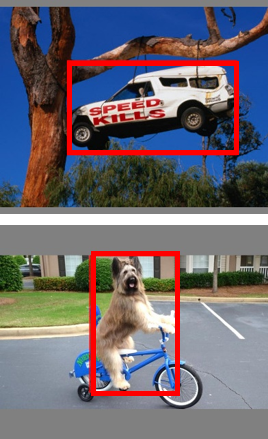}\label{fig:fig2c}}\hspace{0.2cm}
   \subfloat[Schematic of human psychophysics experiment]{\includegraphics[width= 8.5cm]{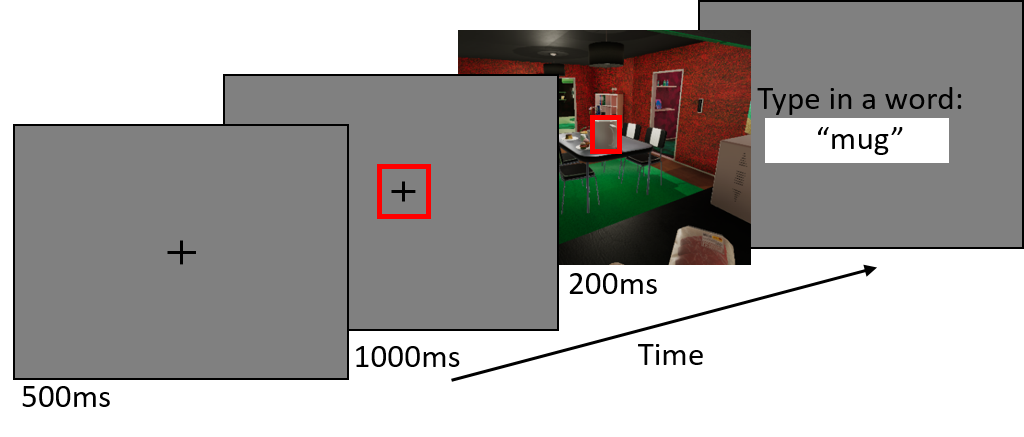}\label{fig:fig2d}}
   \caption{\textbf{Datasets and psychophysics experiment scheme}. (a-c) Example images for each dataset. The red box indicates the target location. In (a), two contextual modifications (gravity and size) are shown.
   In (b), the same target object is cut and pasted into either incongruent or congruent conditions.
   (c) consists of natural images. (d) Subjects were presented with a fixation cross (500 ms), followed by a bounding box indicating the target object location (1000 ms). The image  was shown for 200 ms. After image offset, subjects typed one word to identify the target object.
   }
\label{fig:fig2}\vspace{-4mm}
\end{figure*}

\subsection{Datasets}

\subsubsection{Out-of-context Dataset (OCD)}\label{subsec:VHdataset}

Our out-of-context dataset (OCD) contains 36 object classes, with 
15,773 test images of complex and rich scenes in 6 contextual conditions (described below).
We leveraged the VirtualHome environment \cite{puig2018virtualhome} developed in the Unity simulation engine to synthesize these images in indoor home environments within 7 apartments and 5 rooms per apartment. These rooms include furnished bedrooms, kitchens, study rooms, living rooms and bathrooms \cite{puig2018virtualhome} (see Fig. \ref{fig:fig1} for examples). We extended VirtualHome with additional functionalities to manipulate object properties, such as materials and scales,
and to place objects in out-of-context locations. The target object is always centered in the camera view;  collision checking and camera ray casting are enabled to prevent object collisions and occlusions.

\textbf{Normal Context and No Context:} 
There are 2,309 images with normal context (Fig. \ref{fig:fig1}b), and 2,309 images for the no-context condition (Fig. \ref{fig:fig1}g). 
For the normal context condition, each target object is placed in its ``typical'' location, defined by the default settings of VirtualHome.
We generate a corresponding no context image for every normal context image by replacing all the pixels surrounding the target object with either uniform grey pixels or salt and pepper noise.

\textbf{Gravity:}
We generated 2,934 images where we move the target object along the vertical direction such that it is no longer supported (Fig. \ref{fig:fig1}c). 
To avoid cases where objects are lifted so high that their surroundings
change completely, we 
set the lifting offset to 0.25 meters.


\textbf{Object Co-occurrences:}
To examine the importance of the statistics of object co-occurrences, four human subjects were asked to indicate the most likely rooms and locations for the target objects. We use the output of these responses to generate 1,453 images where we place the target objects on surfaces with lower co-occurrence probability, e.g. a microwave
in the bathroom and Fig. \ref{fig:fig1}d.

\textbf{Object Co-occurrences + Gravity:}
We generated 910 images where the objects are both lifted and placed in unlikely locations. We chose walls, windows, and doorways of rooms where the target object is typically absent (Fig. \ref{fig:fig1}e). We place target objects at half of the apartment's height.

\textbf{Size:} 
We created 5,858 images where we changed the target object's size to 2, 3, or 4 times its original size while keeping the remaining objects in the scene intact (Fig. \ref{fig:fig1}f).
\subsubsection{Real-world Out-of-context Datasets}\label{subsec:realworldocd}

\-\ \textbf{The Cut-and-paste dataset \cite{zhang2020putting}} contains 2,259 out-of-context images spanning 55 object classes. These images are grouped into 16 conditions obtained through the combinations of 4 object sizes and 4 context conditions (normal, minimal, congruent, and incongruent) (Fig. \ref{fig:fig2b}).

\textbf{The UnRel \cite{Peyre17} Dataset} contains more than 1,000 images with unusual relations among objects spanning 100 object classes. The dataset was collected from the web based on triplet queries, such as ``dog rides bike" (Fig. \ref{fig:fig2c}).

\subsection{Performance Evaluation} 

\textbf{Evaluation of Computational Models:} 
We trained the models on natural images from COCO-Stuff \cite{caesar2018coco} using the annotations for object classes overlapping with those in the respective test set (16 overlapping classes between VirtualHome and COCO-Stuff, 55 overlapping classes between Cut-and-paste and COCO-Stuff and 33 overlapping classes between UnRel and COCO-Stuff). 
The models were then tested on OCD, the Cut-and-paste dataset, UnRel, and on a COCO-Stuff test split.


 

\textbf{Behavioral Experiments:} 
We evaluated human recognition on OCD and the Cut-and-paste dataset,
as schematically illustrated in Fig.~\ref{fig:fig2d}, on Amazon Mechanical Turk (MTurk) \cite{turk2012amazon}. We recruited 400 subjects per experiment, yielding $\approx 67,000$ trials. To avoid biases and potential memory effects, we took several precautions: (a) Only one target object from each class was selected;
(b) Each subject saw each room only once;
(c) The trial order was randomized.

Computer vision and most psychophysics experiments enforce N-way categorization (e.g. \cite{tang2018recurrent}). Here we used a more unbiased probing mechanism whereby subjects could use any word to describe the target object. We independently collected ground truth answers for each object in a \emph{separate} MTurk experiment with infinite viewing time and normal context conditions. These Mturk subjects did \emph{not} participate in the main experiments. Answers in the main experiments were then deemed correct if they matched any of the ground truth responses \cite{zhang2020putting}.

A completely fair machine-human comparison is close to impossible since humans have decades of visual+ experience with the world. Despite this caveat, we find it instructive to show results for humans and models on the same images. We tried to mitigate the differences in training by focusing on the qualitative impact of contextual cues in perturbed conditions compared to the normal context condition.
We also show human-model correlations to describe their relative trends across all conditions.

\begin{figure*}
\begin{minipage}{\textwidth}
  \begin{minipage}[b]{11cm}
    \centering
    \includegraphics[width=11cm]{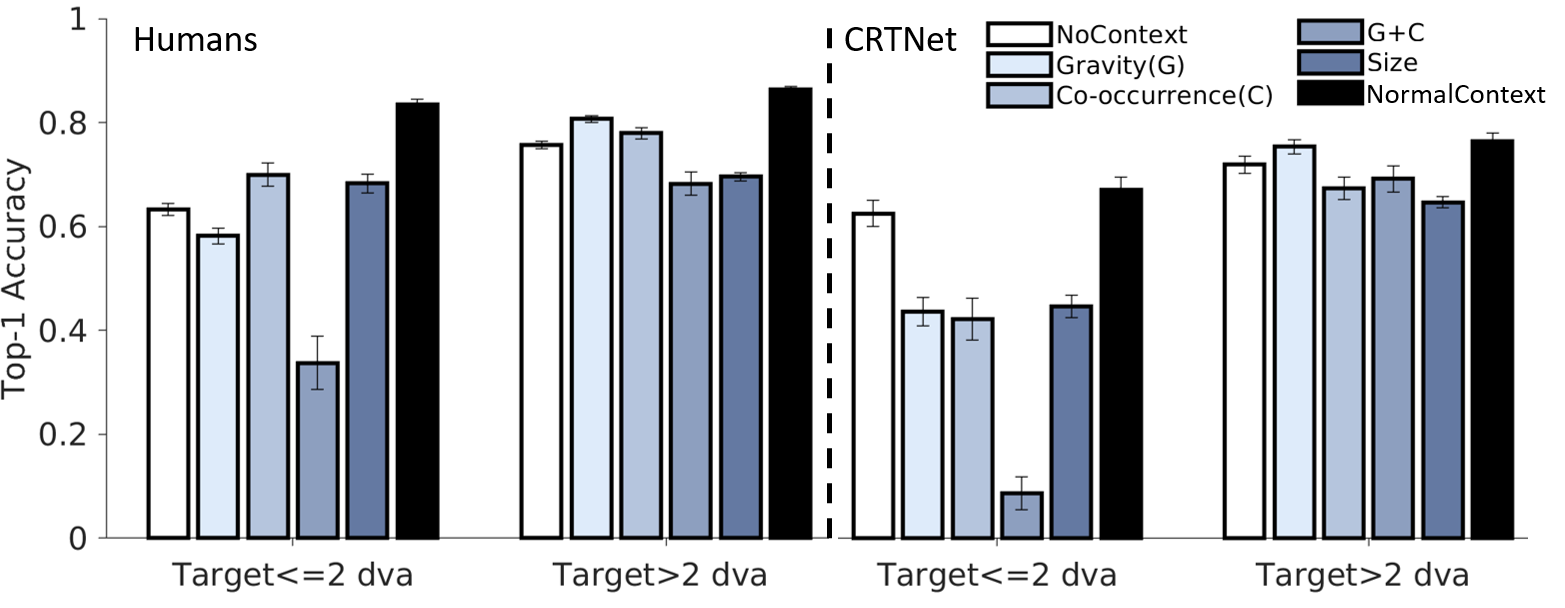}\vspace{-1mm}
    \captionof{figure}{ \textbf{The CRTNet model exhibits human-like recognition patterns across contextual variations in our OCD dataset.}
    Different colors denote contextual conditions (Sec. \ref{subsec:VHdataset}, Fig. \ref{fig:fig1}). We divided the trials into two groups based on target object sizes in degrees of visual angle (dva). 
    Error bars denote the standard error of the mean (SEM).
    }\label{fig:fig4}
  \end{minipage}
  \hfill
  \begin{minipage}[b]{5.5cm}
    \centering
    \begin{tabular}{|c|c|}
\hline
   OCD                                        & Overall \\ \hline
CRTNet (ours)              & \textbf{0.89}      \\ \hline
Baselines & \\ \hline
CATNet \cite{zhang2020putting}  & 0.36       \\ 
Faster R-CNN \cite{ren2015faster} & 0.73       \\ 
DenseNet \cite{huang2017densely} & 0.66       \\ \hline
Ablations & \\ \hline
Ablated-SharedEncoder &0.84      \\ 
Ablated-TargetOnly &\textbf{0.89}      \\ 
Ablated-Unweighted &0.83      \\ 
Ablated-NoDetachment &0.88      \\ \hline
\end{tabular}
      \captionof{table}{\textbf{Linear correlations between human and model performance over 12 contextual conditions.}}\label{tab:tab1}
    \end{minipage}
  \end{minipage}\vspace{-4mm}
\end{figure*}

\section{Results}\label{sec:results}


\subsection{Recognition in our OCD Dataset}\label{subsec:resultsVH}

Figure~\ref{fig:fig4} (left) reports recognition accuracy for humans over the 6 context conditions (Sec. \ref{subsec:VHdataset}, Fig.~\ref{fig:fig1}) and 2 target object sizes (total of 12 conditions). Comparing the no-context condition (white) versus normal context  (black), it is evident that contextual cues lead to improvement in recognition, especially for smaller objects, consistent with previous work ~\cite{zhang2020putting}.
Gravity violations led to a reduction in accuracy. For small objects, the gravity condition was even slightly worse than the no context condition; the unusual context can be misleading for humans.
The effects were similar for the changes in object co-occurrences and relative object size. 
Objects were enlarged by a factor of 2, 3, or 4 in the relative size condition. Since the target object gets larger, and because of the improvement in recognition with object size, we would expect a higher accuracy in the size condition compared to normal context. However, increasing the size of the target object while keeping all other objects intact, violates the basic statistics of expected relative sizes (e.g., we expect a chair to be larger than an apple). Thus, the drop in performance in the size condition is particularly remarkable and shows that violation of contextual cues can override basic object recognition. 

Combining changes in gravity and in the statistics of object co-occcurrences led to a pronounced drop in accuracy. Especially for the small target objects, violation of gravity and statistical co-occurrences led to performance well below that in the no context condition.

These results show that context can play a facilitatory role (compare normal versus no context), but context can also impair performance (compare gravity+co-occurrence versus no context). In other words, unorthodox contextual information hurts recognition.


Figure~\ref{fig:fig4} (right) reports accuracies for CRTNet. Adding normal contextual information (normal context vs no context) led to an improvement of 4\% in performance for both small and large target objects.
Remarkably, the CRTNet model captured qualitatively similar effects of contextual violations as those observed in humans. Even though the model performance was below humans in absolute terms (particularly for small objects), the basic trends associated with the role of contextual cues in humans can also be appreciated in the CRTNet results. Gravity, object co-occurrences, and relative object size changes led to a decrease in performance. As in the behavioral measurements, these effects were more pronounced for the small 
objects. For CRTNet, all conditions led to worse performance than the no context condition for small objects. 


\subsection{Recognition in the Cut-and-paste Dataset}

Synthetic images offer the possibility to systematically control every aspect of the scene, but such artificial images do not follow all the statistics of the natural world. 
Therefore, we further evaluated CRTNet and human performance in the 
naturalistic settings of the Cut-and-paste dataset \cite{zhang2020putting}
(see Table~\ref{tab:tab2}).
The CRTNet model yielded results that were consistent with, and in many conditions better than, human performance. As observed in the human data, performance increases with object size. In addition, the effect of context was more pronounced for smaller objects (compare normal context (NC) versus minimal context (MC) conditions). 

In accordance with previous work \cite{zhang2020putting}, compared to the minimal context condition, congruent contextual information (CG) typically enhanced recognition whereas incongruent context (IG) impaired performance. Although the congruent context typically shares similar correlations between objects and scene properties,  pasting  the  object  in  a  congruent context led to weaker enhancement than the normal context. 
This lower contextual facilitation  may be  due  to  erroneous  relative  sizes between objects, unnatural boundaries created by pasting, or  contextual  cues  specific  to  each  image.   CRTNet  was relatively oblivious to these effects and performance in the congruent condition was closer to that in the normal context condition whereas these differences were more striking for humans.
In   stark   contrast, incongruent   context   consistently degraded recognition performance   below   the   minimal context condition for both CRTNet and humans.  

\subsection{Recognition in the UnRel Dataset}

The Cut-and-paste dataset introduces artifacts (such as unnatural boundaries and erroneous relative sizes) due to the cut-and-paste process. Therefore, we also evaluated CRTNet on the UnRel dataset \cite{Peyre17}. We use the performance on the COCO-Stuff \cite{caesar2018coco} test split as reference for normal context in natural images. CRTNet showed a slightly lower recognition accuracy in the out-of-context setting  (Fig. \ref{fig:fig5}).



\subsection{Comparison with Baseline Models}

\begin{figure}[h]
\vspace{-4mm}
    \begin{center}
        \includegraphics[width=7cm]{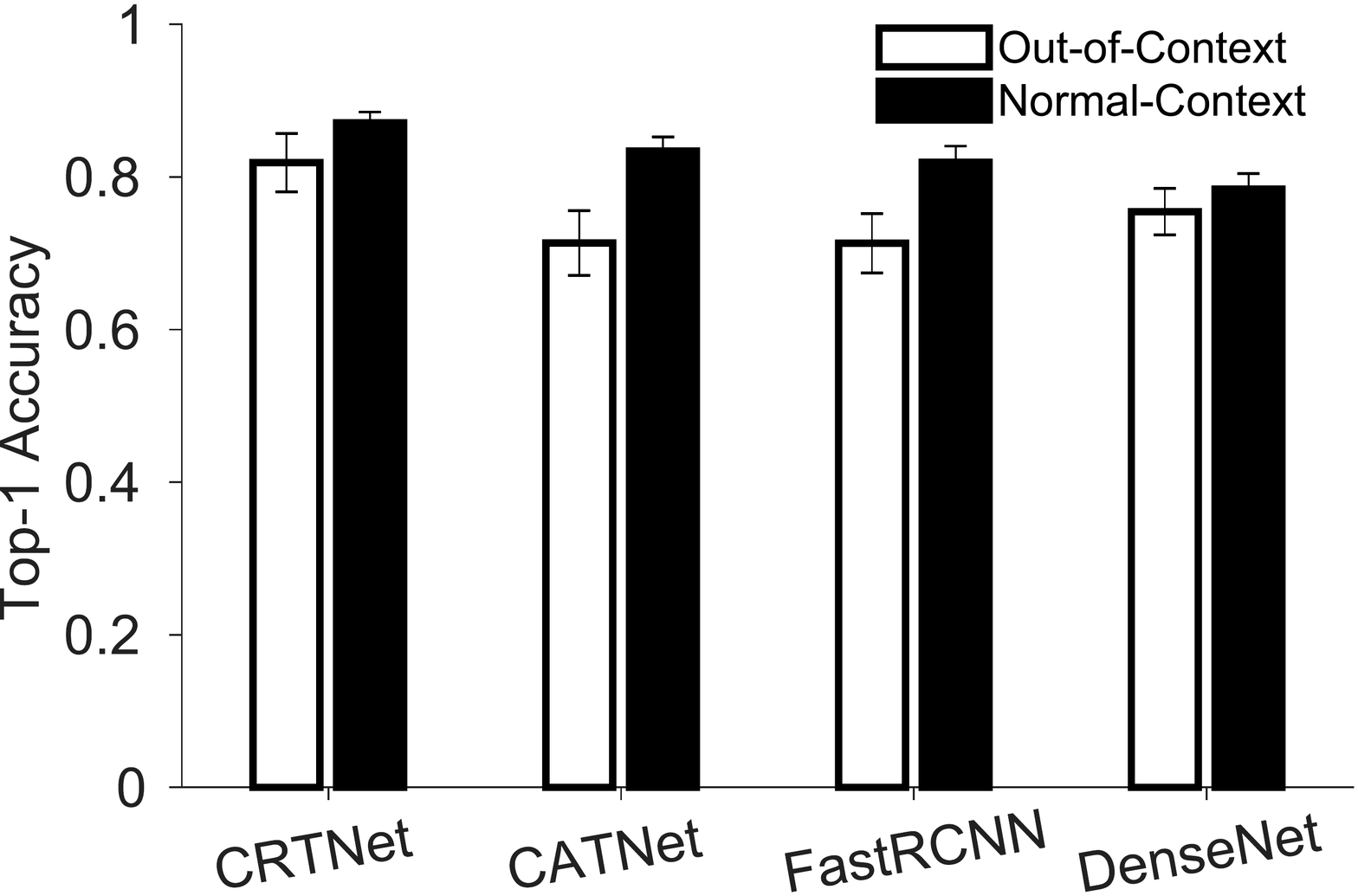}\vspace{-8mm}
    \end{center}
    
    \caption{
    \textbf{CRTNet surpasses all baselines in both normal (COCO-Stuff \cite{caesar2018coco}) and out-of-context (UnRel \cite{Peyre17}) conditions.}
    }
    \label{fig:fig5}
\end{figure}

\textbf{Performance Evaluation:}
Although Faster R-CNN and CATNet leverage global contextual information, CRTNet outperformed both models, especially on small objects (OCD: Fig. \ref{fig:fig4} and Supp. Fig. S7-S8;
Cut-and-Paste: Table\ref{tab:tab2}; UnRel: Fig. \ref{fig:fig5}).
Furthermore, Table \ref{tab:tab1} shows that CRTNet's performance pattern across the different OCD conditions is much more similar to the human performance pattern (in terms of correlations) than the other baseline models.


\begin{table*}[ht]
    \footnotesize
    \centering
    \resizebox{\textwidth}{!}{
    \begin{tabular}{|c|cccc|cccc|cccc|cccc|}
    \hline
             & \multicolumn{4}{c|}{Size {[}0.5, 1{]} dva} & \multicolumn{4}{c|}{Size {[}1.75, 2.25{]} dva} & \multicolumn{4}{c|}{Size {[}3.5, 4.5{]} dva} & \multicolumn{4}{c|}{Size {[}7, 9{]} dva} \\ \hline
             & NC       & CG       & IG       & MC       & NC        & CG        & IG        & MC        & NC        & CG        & IG       & MC       & NC       & CG       & IG      & MC      \\ \hline
    Humans   &\textbf{56.0}&18.8&5.9&10.1&66.8&48.6&22.3&38.9&78.9&66.0&38.8&62.0&88.7&70.7&59.0&77.4 \\
    \cite{zhang2020putting}         &(2.8)&(2.3)&(1.3)&(1.7)&(2.7)&(2.8)&(2.4)&(2.8)&(2.4)&(2.7)&(2.6)&(2.8)&(1.7)&(2.6)&(2.8)&(2.3)\\ \hline
    CRTNet      & 50.2&\textbf{43.9}&10.6&\textbf{17.4}&\textbf{78.4}&\textbf{81.4}&\textbf{41.2}&\textbf{56.7}&\textbf{91.5}&\textbf{87.3}&51.1&\textbf{76.6}&\textbf{92.9}&\textbf{87.7}&\textbf{66.4}&\textbf{83.0}\\
     (ours)        &  (2.8)&(2.8)&(1.7)&(2.1)&(3.0)&(2.8)&(3.5)&(3.6)&(1.1)&(1.3)&(1.9)&(1.6)&(0.9)&(1.2)&(1.7)&(1.4) \\ \hline
    CATNet   &37.5&29.2&3.6&6.1&53.0&46.5&10.9&22.1&72.8&71.2&24.5&38.9&81.8&78.9&47.6&74.8\\
    \cite{zhang2020putting}         &(4.0)&(2.4)&(1.0)&(2.0)&(4.1)&(2.5)&(1.6)&(3.6)&(3.6)&(2.4)&(2.2)&(3.9)&(3.0)&(2.1)&(2.6)&(3.5)\\ \hline
    Faster R-CNN &24.9&10.9&5.9&7.2&44.3&27.3&20.1&16.5&65.1&53.2&39.0&42.9&71.5&64.3&55.0&64.6\\
      \cite{ren2015faster}       &(2.4)&(1.7)&(1.3)&(1.4)&(3.6)&(3.2)&(2.9)&(2.7)&(1.8)&(1.9)&(1.9)&(1.9)&(1.6)&(1.7)&(1.8)&(1.7)\\ \hline
    DenseNet &13.1&10.0&\textbf{11.2}&12.5&45.4&42.3&39.7&46.4&67.1&62.3&\textbf{55.4}&67.1&74.9&67.2&63.5&74.9\\
     \cite{huang2017densely}        &(1.9)&(1.7)&(1.8)&(1.8)&(3.6)&(3.5)&(3.5)&(3.6)&(1.8)&(1.9)&(1.9)&(1.8)&(1.6)&(1.7)&(1.7)&(1.6)\\ \hline
    \end{tabular}
    }
    \caption{\textbf{Recognition accuracy of humans, the CRTNet model, and three different baselines on the Cut-and-paste dataset \cite{zhang2020putting}}. There are 4 conditions for each size: normal context (NC), congruent context (GC), incongruent context (IG) and minimal context (MC) (Sec. \ref{subsec:realworldocd}). Bold highlights the best performance.
    Numbers in brackets denote the standard error of the mean.
    }
    \vspace{-4mm}
    \label{tab:tab2}
\end{table*}

\textbf{Architectural Differences:}
While all baseline models can rely on an intrinsic notion of spatial relations, CRTNet learns about spatial relations between target and context tokens through a positional embedding.
A visualization of the learned positional embeddings (Supp. Fig. S1) shows that
CRTNet learns image topology by encoding distances within the image in the similarity of positional embeddings.


In CATNet, the attention map iteratively modulates the extracted feature maps from the context image at each time step in a recurrent neural network,
whereas CRTNet uses a stack of feedforward transformer decoder layers with multi-head encoder-decoder attention. These decoder layers hierarchically integrate information via attention maps, modulating the target token features with context. 



DenseNet takes cropped targets as input with only a few surrounding pixels of context.
Its performance dramatically decreases for smaller objects, which also results in lower correlation with the human performance patterns. For example, in the Cut-and-paste dataset, CRTNet outperforms DenseNet by 30\% for normal context and small objects (Table ~\ref{tab:tab2}) and in OCD, DenseNet achieves a correlation of 0.66 vs. 0.89 for CRTNet (Table~\ref{tab:tab1}).

\subsection{Ablation Reveals Critical Model Components}\label{subsec:ablation}

We assessed the importance of design choices by training and  testing  ablated  versions  of  CRTNet on the OCD dataset. 

\textbf{Shared Encoder:} In the CRTNet model, we trained two separate encoders
to extract features from target objects and the context respectively. Here, we enforced weight-sharing between these two encoders (Ablated-SharedEncoder) to assess whether the same features for both streams are sufficient to reason about context.
The results (Table~\ref{tab:tab1}, Supp. Fig. S3) show that
the ablated version achieved a lower recognition accuracy and lower correlation with the psychophysics results.

\textbf{Recognition Based on Target or Context Alone:} 
In the original CRTNet model, we use the confidence-weighted prediction $y_p$.
Here, we tested two alternatives: CRTNet relying only on the target object ($y_t$, Ablated-TargetOnly) and CRTNet relying only on contextual reasoning ($y_{t,c}$, Ablated-Unweighted). 
The original model benefits from proper contextual information compared to the target-only version but it is slightly more vulnerable to some of the context perturbations as would be expected. It consistently outperforms the context-only version demonstrating the usefulness of the confidence-modulation mechanism.

\textbf{Joint Training of the Target Encoder:} In Sec. \ref{subsec:trainingtest}, we use gradient detachments to make the training of the target encoder $E_t(\cdot)$ independent of $G_t(\cdot)$ such that it cannot force the target encoder to learn more discriminative features. 
Here we remove this constraint (Ablated-NoDetachment,  Supp. Fig. S6).
The results are inferior to the ones of our original CRTNet, supporting the use of the gradient detachment method.


\vspace{-2mm}
\section{Conclusion}
We introduced the OCD dataset and used it to systematically and quantitatively study
the role of context
in object recognition. OCD allowed us to rigorously scrutinize the multi-faceted aspects of how contextual cues influence visual recognition.
We conducted experiments with computational models and complemented them with psychophysics studies to gauge human performance.
Since the synthetic images in OCD can still be easily distinguished from real photographs, we addressed potential concerns due to the domain gap with experiments on two additional datasets consisting of real-world images.

We showed consistent results for humans and computational models over all three datasets. The results demonstrate that contextual cues can enhance visual recognition, but also that the ``wrong'' context can impair visual recognition capabilities both for humans and models.

We proposed the CRTNet model as a powerful and robust method to make use of contextual information in computer vision. CRTNet performs well compared to competitive baselines across a wide range of context conditions and datasets. In addition to its performance in terms of recognition accuracy, CRTNet's performance pattern was also found to resemble human behavior more than that of any baseline model.



\vspace{1mm}
{\small
\noindent \textbf{Acknowledgements}
This work was supported by NIH R01EY026025 and the Center for Brains, Minds and Machines, funded by NSF STC award CCF-1231216. MZ is supported by a postdoctoral fellowship of the Agency for Science, Technology and Research. 
We thank Leonard Tang, Jeremy Schwartz, Seth Alter, Xavier Puig, Hanspeter Pfister, Jen Jen Chung, and Cesar Cadena for useful discussions and support.
}

{\small
\bibliographystyle{ieee_fullname}
\bibliography{egbib}
}

\end{document}